\documentclass[letterpaper, 10 pt, conference]{ieeeconf}  

\IEEEoverridecommandlockouts                              

\usepackage{graphics} 
\usepackage{epsfig} 
\usepackage{mathptmx} 
\usepackage{times} 
\usepackage{amsmath} 
\usepackage{amssymb}  

\usepackage{algorithmic}

\usepackage{amsmath,amsfonts}
\usepackage{array}
\usepackage[caption=false,font=normalsize,labelfont=sf,textfont=sf]{subfig}
\usepackage{textcomp}
\usepackage{stfloats}
\usepackage{url}
\usepackage{verbatim}
\usepackage{graphicx}

\usepackage[font=footnotesize,labelfont=bf]{caption}
\usepackage{hyperref}
\usepackage{float}
\makeatletter
\let\newfloat\newfloat@ltx
\makeatother

\usepackage[ruled,vlined]{algorithm2e}
\usepackage{amssymb}  
\usepackage{color}
\usepackage{colortbl}
\usepackage[dvipsnames]{xcolor}
\usepackage{mathtools}
\usepackage[utf8]{inputenc}
\usepackage{xr}
\usepackage{mathrsfs}
\usepackage{enumerate}
\usepackage{multicol}
\usepackage{cite}
\usepackage{changepage}
\usepackage{mdframed}

\usepackage{wrapfig}

\usepackage{booktabs}

\usepackage{tabularx}

\usepackage{booktabs}
\usepackage[normalem]{ulem}
\useunder{\uline}{\ul}{}

\usepackage[textsize=tiny]{todonotes}

\title{\LARGE \bf
SAFE-GIL: SAFEty Guided Imitation Learning for Robotic Systems
\vspace{-0.5em}
}
\author{Yusuf Umut Ciftci, Darren Chiu, Zeyuan Feng, Gaurav S. Sukhatme, Somil Bansal
\thanks{$^{1}$Authors are with ECE at the University of Southern California, LA, USA. Corresponding author: {yciftci@usc.edu}.}%
}

\renewcommand{\baselinestretch}{0.975}

\begin{document}

\maketitle
\thispagestyle{empty}
\pagestyle{empty}


\begin{abstract}
Behavior cloning (BC) is a widely-used approach in imitation learning, where a robot learns a control policy by observing an expert supervisor. 
However, the learned policy can make errors and might lead to safety violations, which limits their utility in safety-critical robotics applications.
While prior works have tried improving a BC policy via additional real or synthetic action labels, adversarial training, or runtime filtering, none of them explicitly focus on reducing the BC policy's safety violations during training time.
We propose SAFE-GIL, a design-time method to learn safety-aware behavior cloning policies.
SAFE-GIL deliberately injects adversarial disturbance in the system during data collection to guide the expert towards safety-critical states. 
This disturbance injection simulates potential policy errors that the system might encounter during the test time.
By ensuring that training more closely replicates expert behavior in safety-critical states, our approach results in safer policies despite policy errors during the test time.
We further develop a reachability-based method to compute this adversarial disturbance.
We compare SAFE-GIL with various behavior cloning techniques and online safety-filtering methods in three domains: autonomous ground navigation, aircraft taxiing, and aerial navigation on a quadrotor testbed.
Our method demonstrates a significant reduction in safety failures, particularly in low data regimes where the likelihood of learning errors, and therefore safety violations, is higher.
See our website here: \url{https://y-u-c.github.io/safegil/}
\end{abstract}


\section{Introduction} \label{Introduction}
Imitation learning offers a powerful framework for teaching complex behaviors to robots without requiring detailed reward signals. 
A key technique in this domain is Behavior Cloning (BC), where a robot learns direct mappings from states or observations to actions by emulating a supervisor \cite{alvin, il_survey}.
BC has been used across a variety of robotic applications, ranging from manipulation \cite{il_manipulation1, il_manipulation_review}, navigation \cite{bansal2019combining}, to autonomous driving \cite{cahuffernet, alvin, condinional_il, agile_auto_driving}.
However, in general, even when the expert demonstrations themselves are safe, it is hard to ensure the safety of the resultant policy. 
This can happen, for instance, due to potential learning errors or distribution shifts due to the closed-loop nature of control (sometimes also referred to as covariate shift), limiting the use of BC in safety-critical robotics applications.
In this work, our goal is to learn robust BC policies to mitigate safety-critical policy failures.

\begin{figure}[h!]
\centering
\includegraphics[width=0.49\textwidth]{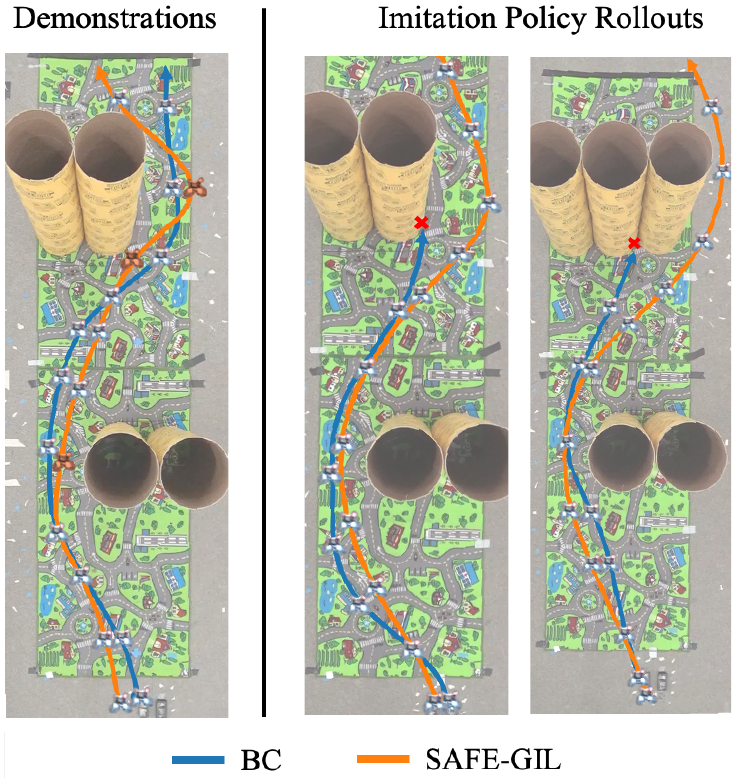}
\caption{Left: Human controlled quadrotor demonstration trajectories with (SAFE-GIL) and without adversarial safety guidance. With SAFE-GIL, the robot observes more safety-critical states during training time (illustrated by red quadrotor icons). Right: BC and SAFE-GIL policy rollouts. The red cross denotes a collision. SAFE-GIL results in a significant improvement in robot safety during the test time.}
\label{fig:dubins_main}
\vspace{-2.5em}
\end{figure}

Several works have studied the safety problem under BC policies.
These methods can broadly be classified into training time (or \textit{design-time}) methods and test time (or \textit{deployment-time}) methods.
A popular test-time approach is safety filtering, wherein a recovery policy/mechanism is used whenever the BC policy is at risk of violating the safety constraint \cite{reichlin2022back, wabersich2023data, hsu2023safety, yang2024enhancing}.
A key advantage of the safety filtering approach is its simplicity -- it is agnostic to how the base policy was trained.
Correspondingly, a number of methods have been developed to construct safety filters, such as Control Barrier functions (CBF) \cite{ames2017control}, Hamilton-Jacobi (HJ) reachability \cite{bajcsy2019efficient, borquez2023safety}, Model Predictive Control \cite{wabersich2023data}, or learning from demonstrations \cite{yang2024enhancing}.
However, safety filtering has a few limitations in the context of imitation learning -- despite recent progress, it remains challenging to construct and update filters online for autonomous systems, especially when the system state is not available (e.g., when the robot is operating based on high-dimensional sensory information). 
Furthermore, the overall performance of the filtered policy is still limited by the underlying BC policy; therefore, it is desirable to \textit{design} BC policies that proactively reason about system safety, resulting in overall performance improvement.

To that end, design-time methods have also been studied. \cite{cosner2022end} uses CBF-based filtering to modify the expert controller itself to enhance the safety of the learned policy. However, a direct modification of the expert controller can interfere with the task at hand.
Other design-time methods focus on modifying the data collection procedure or augmenting the expert data, with the overall goal of reducing learning error and covariate shift, which, in turn, can reduce safety violations \cite{totsila2023end, khansari2011learning, mehta2024strol, pfrommer2022tasil, kang2022lyapunov}. 
For instance, off-policy methods perturb the expert to collect more diverse examples \cite{laskey2017dart} or synthetically augment the expert demonstrations to gather a larger dataset \cite{ke2023ccil, ke2021grasping, zhou2023nerf, chang2021mitigating, hoque2024intervengen}.
Adversarial imitation learning methods \cite{gail, airl}, interact with the environment to determine actions that bring a policy’s occupancy measure closer to the expert’s by training the policy adversarial to a discriminator, whereas on-policy methods add such actions to the robot's dataset as the robot executes what it has learned in the environment \cite{ross2011reduction, hoque2021thriftydagger, kelly2019hg, spencer2022expert, mehta2023unified}. 
Other methods define safety as the deviation from the expert controller\cite{zhang2016query, menda2019ensembledagger} and query expert whenever a deviation threshold is reached.
However, this perspective on safety is different from satisfying a constraint or avoiding a failure set (e.g., avoiding collision with obstacles or not entering a no-go zone). 

To overcome these challenges, we propose SAFE-GIL
(\underline{SAFE}ty \underline{G}uided \underline{I}mitation \underline{L}earning) -- a novel \textit{design-time} algorithm to learn safety-aware BC policies. 
Our key insight is to inject adversarial disturbance during data collection to simulate potential test-time safety-critical learning errors. 
At any state, the added disturbance corresponds to the most safety-critical policy error, meaning the resultant robot state is closest to violating safety constraints. 
By injecting such disturbances into expert demonstrations, we intentionally steer the system toward riskier states and learn corrective actions. 
Our hypothesis is that collecting expert demonstrations in these states will enable the system to recover from them should any safety-critical error occur during the test time, resulting in safer policies.
We also propose a mechanism to automatically synthesize such disturbances via offline reachability-based computations.
Overall, SAFE-GIL inherits the advantages of BC methods while prioritizing the enhancement of system safety under the learned policy. 

To summarize, our key contributions are: (a) we propose a new algorithm, SAFE-GIL, to learn safety-aware imitation policies to reduce test-time safety failures; (b) through simulation studies and hardware experiments on a quadrotor testbed, we demonstrate that SAFE-GIL results in significantly safer imitation policies, while maintaining system's performance; (c) we demonstrate that SAFE-GIL can be seamlessly integrated with existing imitation learning approaches, such as DAgger, to reduce covariate shift while mitigating safety-critical failures. Additionally, it can be combined with online safety mechanisms, such as safety filtering, to further reduce test-time failures.
\vspace{-0.7em}
\section{Problem Formulation} \label{sec:problem_setup}
Consider a robot with state $x \in X \subseteq \mathbb{R}^n$ and control $u \in \mathcal{U}$.
We also assume access to an approximate dynamics model of the robot $\dot{x}=f(x, u)$ during training.
Let $\pi^*(\cdot)$ denote an expert policy that we want to replicate via an imitation policy $\pi_{\theta}(\cdot)$, where $\theta \in \Theta$ is to be learned.
Finally, let $L \subseteq \mathbb{R}^n$ be the set of states that represent failure for the robot.
For example, $L$ might represent obstacles for a navigation robot or off-runway positions for an autonomous aircraft.

\noindent \textit{\textbf{Objective.}} Our goal in this work is to learn an imitation policy that minimizes safety violations (i.e., stay outside $L$) during the test time, while minimally sacrificing the policy performance.
\section{Background: Hamilton-Jacobi Reachability}
We now provide a brief overview of Hamilton-Jacobi (HJ) reachability analysis which we will use to compute adversarial disturbances to guide the expert towards safety-critical states.
HJ reachability analysis is a formal verification method for characterizing the safety-critical regions of a dynamical system \cite{bansal2017hamilton, mitchell2005time}.
For reachability analysis, we will consider a more general form of dynamics: $\dot{x}=f(x, u, d)$, where $d \in [-\bar{d}, \bar{d}]$ represents the disturbance or uncertainty in the dynamics.
Later on, in our work, we will use $d$ to model potential policy errors.
Given the dynamics $f$ and a failure set \( L \), we are interested in computing the backward reachable tube (BRT) of the system -- the set of all states from which the system will eventually enter the failure set regardless of the control policy $u(\cdot)$, given the system is subject to optimum disturbance policy $d(\cdot)$ \cite{bansal2017hamilton}.

To compute the BRT, we represent the failure set $L$ implicitly with a target function \( l(x) \), i.e., \( L = \{ x : l(x) \leq 0 \} \).
Next, we define the cost function as the minimum value of the target function along the robot trajectory, starting from a state $x_0$ at time $t$:
\begin{equation}
\label{safety_cost}
\vspace{-0.5em}
J(x_0, t, u(\cdot), d(\cdot)) = \min_{\tau\in[t,T]} l\left (
x(\tau)\right ). \quad 
\end{equation}
Thus, the sign of the cost function tells us whether a robot trajectory ever entered the failure set.
Finally, the optimal $u(\cdot)$ drives the system away from the unsafe states (i.e., it maximizes the above cost function) and the optimal $d(\cdot)$ drives the system towards $L$ (i.e., minimizes the cost function). 
The robust value function can be obtained using the principle of dynamic programming, which results in solving the following final value Hamilton-Jacobi-Isaacs Variational Inequality (HJI-VI).
\begin{equation}
\label{hji-vi}
\min \left\{ D_t V(x,t) + H(x,t), l(x) - V(x,t) \right\} = 0; \quad  V(x,T) = l(x),
\end{equation}
where \( D_t \) and \( \nabla \) represent the time and spatial gradients of the value function and \( H \) is the Hamiltonian given as:
\begin{equation}
\vspace{-0.5em}
H(x, t) = \max_u \min_d \langle \nabla V(x, t), f(x, u, d) \rangle. \quad 
\end{equation}
For a detailed derivation and discussion of the HJI-VI in \eqref{hji-vi}, we refer the interested readers to \cite{mitchell2005time} and \cite{bansal2017hamilton}. 
Intuitively, the value function at $x$ represents the closest the robot will ever get to the failure set under the optimal safety policy starting from state $x$.
Thus, a lower value denotes a more safety-critical state, and a value below 0 denotes an unrecoverable state, i.e., the state is in the BRT.

The value function typically converges after some time horizon, beyond which the system has enough time to maneuver to avoid the failure set despite the worst case disturbance. 
Consequently, we can use the converged value function, $V^*(x) = \lim_{t \rightarrow \infty} V(x, t)$.
Given $V^*$, the BRT is given as the sub-zero level set of the value function: $\mathcal{V} = \{x : V^*(x) \leq 0\}$.
Importantly, the value function can be used to synthesize the optimal disturbance that at any state $x$ maximally steers the robot toward unsafe states:
\begin{equation}
\label{eq:opt_dist}
d^*(x) = \max_u \arg \min_d \langle \nabla V^*(x), f(x, u, d) \rangle,
\end{equation}
i.e., towards lower values.
We will use $d^*$ to guide the expert towards safety-critical states.

\noindent \textbf{HJ Reachability Computation.}
Several methods have been proposed to solve the HJI-VI in \eqref{hji-vi} and compute the value function (see \cite{bansal2017hamilton} for a survey).
One method is to numerically solve the HJI-VI over a discretized grid in the state space \cite{mitchell2004toolbox, bansal2020provably}, which is what we use in this work to compute the value function for the simulation case studies. Other learning-based methods also exist that are more scalable to high-dimensional systems, such as DeepReach \cite{deepreach}, which we use for the quadrotor hardware experiment.
\section{SAFE-GIL: SAFEty Guided Imitation Learning}
We aim to learn safe imitation policies by intentionally guiding the expert toward safety-critical states during data collection. 
Our key idea is to abstract potential test-time policy errors as an \textit{adversarial disturbance} that attempts to steer the agent towards unsafe states over time, and inject this disturbance during data collection.
Specifically, the expert is guided towards safety-critical states by the application of a perturbed input, $\pi^G(x)$, to the system, instead of $\pi^*(x)$:
\begin{equation} \label{eqn:dist_addition}
    \pi^G(x) = \pi^*(x) + d(x),
\end{equation}
where $d(x)$ is the injected disturbance.
Intuitively, $d(x)$ can be thought of as simulating potential learning error at state $x$ during the test time. 
However, since the learning error is not known beforehand, we inject $d(x)$ so as to maximally steer the system towards the failure set.
Overall, by guiding the expert with $\pi^G$, the system is likely to visit safety-critical states more often.
The corrective examples from these states will allow the robot to learn to recover from them, should they be encountered during the test time, enhancing the overall system safety.

\vspace{-1em}
\begin{algorithm}[h!]
    \caption{Safety Guided Data Collection}
    \label{our_algorithm}
\begin{algorithmic}
\STATE \textbf{Input:} Number of demonstrations K; Trajectory length T\;
\STATE \textbf{Output:} Guided state trajectory and corresponding expert actions $(x, \pi^*(x))$\;
\FOR{i=1:K}
    \STATE Sample initial state $x_1$
    \FOR{j=1:T}
        \STATE $\bar{d} \sim U(0,\bar{d}_{max})$
        \STATE Get optimal disturbance $d^*(x_i; \bar{d})$ using \eqref{eq:opt_dist}
        \STATE $\pi^G(x_i) \gets \pi^*(x_i) + d^*(x_i; \bar{d})$,
        \STATE Store $(x_i, \pi^*(x_i))$ pair
        \STATE Apply $\pi^G(x_i)$ on the system; obtain $x_{i+1}$
    \ENDFOR
\ENDFOR
\STATE \textbf{return} State and expert action pairs
\end{algorithmic}
\end{algorithm}
\vspace{-1em}

\noindent \textbf{Computation of $d(x)$.}
In this work, we use HJ reachability analysis to compute $d(x)$. 
Specifically, given the disturbance injection scheme in \eqref{eqn:dist_addition}, the robot state evolution can approximately be described as $\dot{x}=f(x, u + d) := g(x, u, d)$, where $d \in [-\bar{d}, \bar{d}]$.
$\bar{d}$ is a hyperparameter that can be used to control the amount of injected disturbance.
Given the disturbance-injected dynamics, $g$, and the failure set $L$, we compute the corresponding safety value function $V^*(x)$ by solving the HJI-VI in \eqref{hji-vi}.
To compute the value function for different disturbance bounds, we further condition the value function on $\bar{d}$ to obtain a parameter-conditioned value function $V^*(x; \bar{d})$.
The parameter-conditioned value function can be obtained by simply adding a dummy state $\bar{d}$ to the system with zero dynamics \cite{parameter_conditioned_brt}.
Given the value function, the optimal adversarial disturbance $d^*(x; \bar{d})$ is obtained using \eqref{eq:opt_dist}, which steers the system towards the smallest $V^*$ at any state, i.e., towards more safety-critical states.

Finally, at each time step during data collection, the upper bound on disturbance, $\bar{d}$, is uniformly sampled between $[0, \bar{d}_{max}]$, i.e., $\bar{d} \sim U(0,\bar{d}_{max})$.
Correspondingly, we obtain the optimal adversarial disturbance $d^*(x; \bar{d})$.
Scaling of the disturbance bound helps in various ways: first, injecting disturbance with the maximum bound at each time step can steer the system too close to the failure set during data collection, which is not desirable for safety-critical systems. 
Second, since we do not know the policy error distribution beforehand, randomizing the disturbance magnitude results in more diverse demonstrations without being overly conservative.
Our overall data collection procedure is described in Algorithm \ref{our_algorithm}.
To collect a demonstration trajectory, we randomly sample an initial state. Next, at each time step along the trajectory, we uniformly sample a $\bar{d}$ and compute the optimal disturbance using $V^*(x; \bar{d})$ and \eqref{eq:opt_dist}. The disturbance is then injected into the expert action and applied to the system, and the entire process is repeated until the end of the trajectory.
Finally, we train a behavior cloning policy $\pi_{\theta}(x)$ on the collected data. 
\section{Experiments} \label{sec:experiments}
We demonstrate the robustness of the learned policy under SAFE-GIL on two simulation case studies (state-based autonomous navigation and camera-based aircraft taxiing) and on a hardware testbed (aerial navigation). 
Each study varies in dynamics, observation space, and compute resources, with the intention of demonstrating safety enhancement.
\subsection{Autonomous Navigation Using a State-Based Policy}
\label{dubins_ex}
We consider a wheeled robot navigating in a 2D space to reach a goal position without colliding with obstacles in the environment. 
The navigation task is to be performed autonomously during the test time, starting from various initial states.
The robot state is given by $x := (p_x, p_y, \theta)$, where $(p_x, p_y)$ is the robot position and $\theta$ is its heading. 
The control input is angular velocity $u := \omega$, bounded by $|\omega| \leq \bar{\omega}$.
We model the robot as a unicycle with dynamics:
$\dot{p_x} = v \cos(\theta), \quad \dot{p_y} = v \sin(\theta),\quad \dot{\theta} = \omega$.
In our example, we use $\bar{\omega} = 1 \, rad/s$ and $v = 1 \, m/s$.

The failure set $L$ is given by the gray obstacles (Fig. \ref{fig:dubins_main}) that the robot must avoid to reach the goal (green area).
The expert is a Model Predictive Control (MPC)-based controller that minimizes a cost function penalizing the robot's distance to the goal, obstacle penetration, and control energy.
The imitation agent takes as input the current robot state $(p_x, p_y, \theta)$ and outputs the angular speed, and is modeled as an MLP.
Each imitator is trained for 10 different seeds to capture the performance variance.

\begin{figure}[h!]
\vspace{-1em}
\centering
\includegraphics[width=0.49\textwidth]{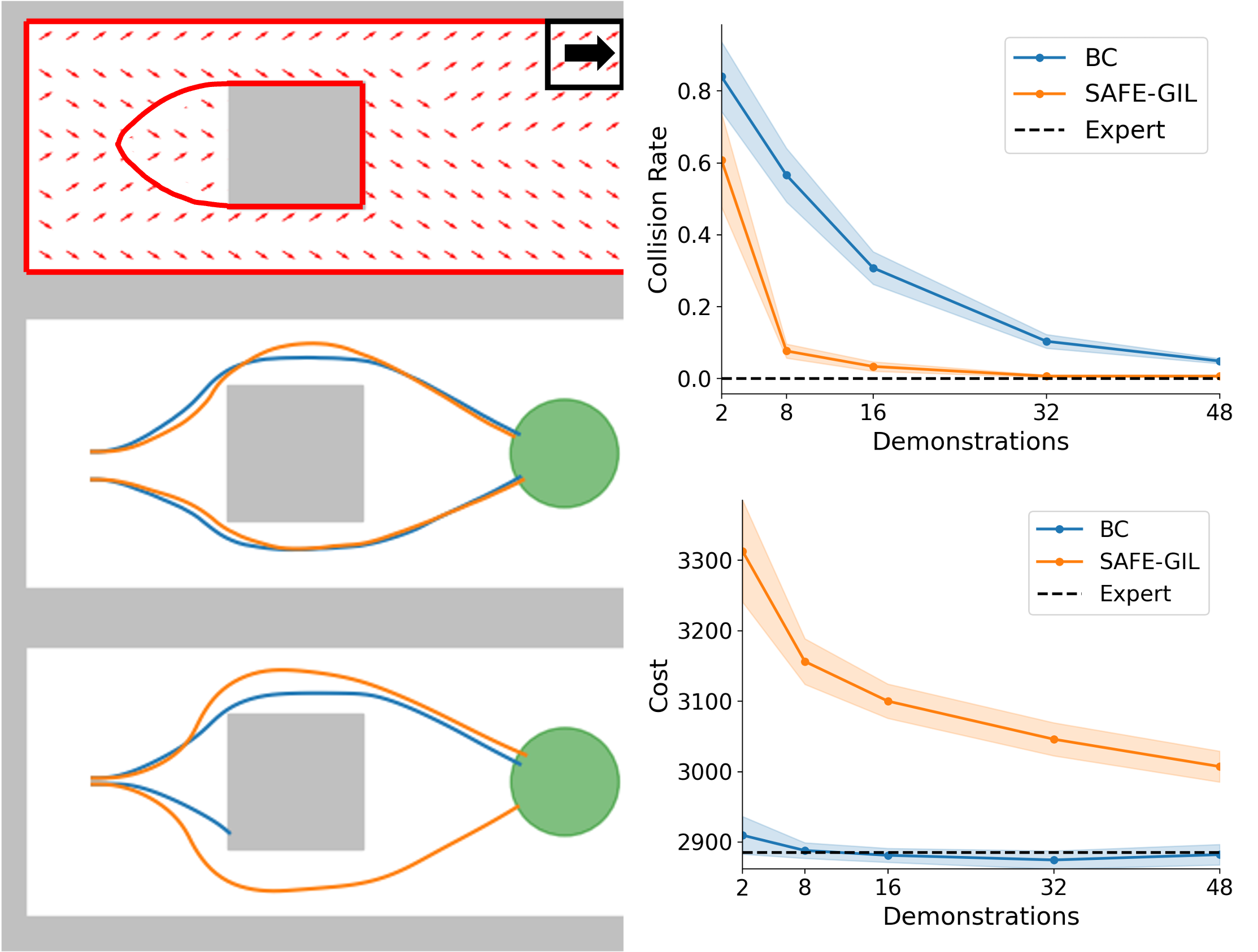}
\caption{Top row: Computed BRT and disturbance for $\theta = 0$. Middle row: Demonstration trajectories with (Orange) and without (Blue) disturbance injection. Bottom row: BC and SAFE-GIL policy rollouts. Right column (Top): Mean collision rate and (Bottom) cost of safe trajectories vs number of demonstrations. SAFE-GIL results in a significant safety improvement.}
\label{fig:dubins_main}
\vspace{-1em}
\end{figure}

For SAFE-GIL, we use $\bar{d}_{max} = 0.6~\bar{\omega}$, which corresponds to a maximum of $60\%$ policy error in imitating the agent.
The value function is computed over a $101 \times 101 \times 101$ grid using the Level Set Toolbox \cite{level_set_toolbox}
by solving the HJI-VI numerically (converging within $\approx 17s$).
The converged value function is then used to compute $d^*$ for guidance.
In Fig. \ref{fig:dubins_main} (top), the red region indicates the BRT -- if the robot enters this set, it doesn't have enough control authority (angular velocity) to avoid a collision with the obstacle.
The red arrows indicate the direction of $d^*$ at each state,
which pushes the robot towards the obstacles.

\noindent \textbf{\textit{Safety.}}
We compare SAFE-GIL against vanilla Behavior Cloning (BC) for different number of expert demonstrations.
Fig. \ref{fig:dubins_main} compares the collision rates (percentage of trajectories that collide before reaching the goal location) of the resulting imitation policies by rolling out from 100 randomly sampled initial states.
SAFE-GIL is able to achieve substantially lower collision rates compared to BC, especially in low data regimes where policy errors are more likely and the safety problem is more pronounced.
To understand this behavior, we illustrate the expert demonstrations for BC and SAFE-GIL (middle row), as well as the corresponding rollouts of the imitation policies (bottom row) in Fig. \ref{fig:dubins_main}.
Due to disturbance injection, the expert under SAFE-GIL is guided closer to the obstacles; consequently, expert recovery maneuvers from these states are present in the demonstration data, allowing the learned imitation policy to better avoid the obstacles. 
On the other hand, without the proposed disturbance guidance, some of these safety-critical states are never encountered by the expert, even as the number of collected expert trajectories increases, resulting in a slower safety improvement.

\noindent \textbf{\textit{Performance tradeoff}.}
Under SAFE-GIL, the training distribution shifts towards more safety-critical states.
This might result in having fewer samples from the states that maximize the expert's reward function, ultimately leading to performance degradation compared to BC. To distinguish between safety and task performance, we plot the mean cost accrued per safe trajectory of policy rollouts in Fig. \ref{fig:dubins_main}, using the cost function expert is minimizing for. Note that although SAFE-GIL results in more number of safe trajectories compared to BC, the performance of the safe trajectories is lower.

\begin{figure}[ht]
\vspace{-1em}
\centering
\includegraphics[width=0.5\textwidth]{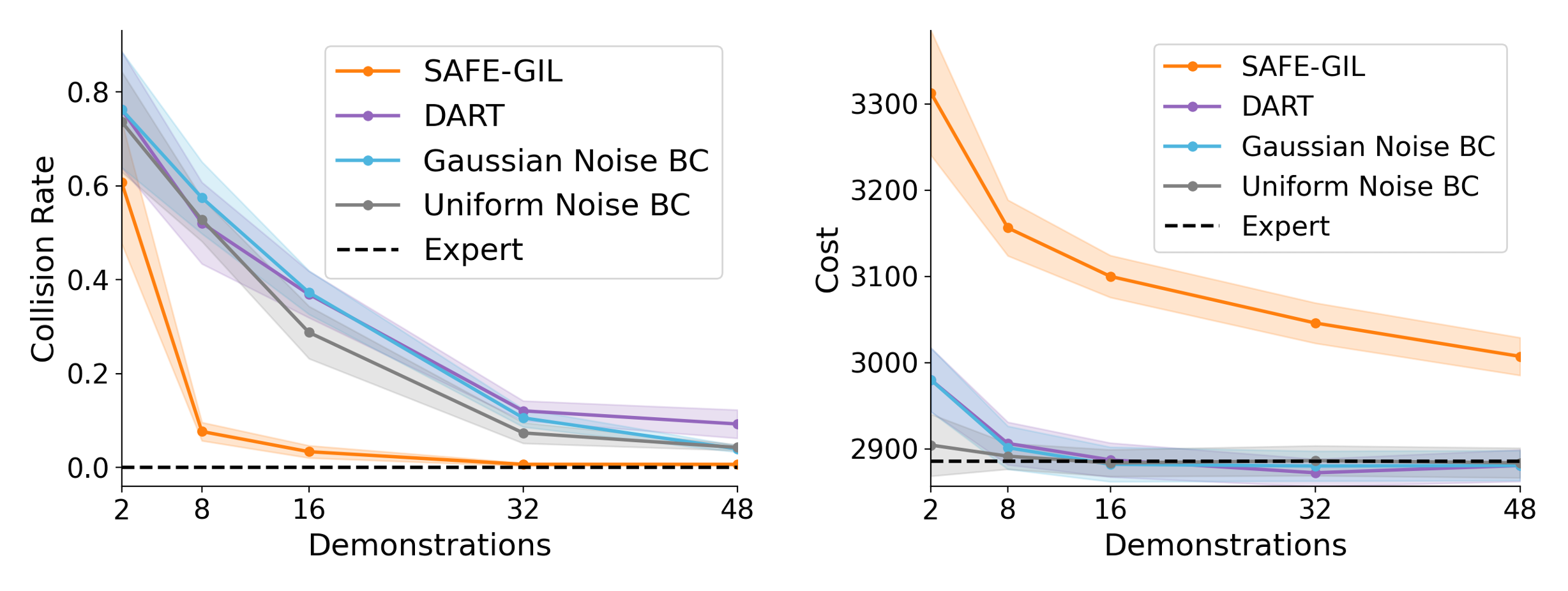}
\caption{Mean collision rate (Left) and cost of safe rollouts (Right) vs number of demonstrations. Adversarial noise injection leads to a significant safety improvement over random noise.}
\label{fig:dubins_noise}
\vspace{-1em}
\end{figure}

\noindent \textbf{\textit{Effect of adversarial disturbance}.}
SAFE-GIL can also be thought of as a data augmentation scheme that adds noise to the expert data. 
To understand the importance of using \textit{adversarial} noise during data augmentation, we compare against other schemes that inject Gaussian noise (\textbf{Gaussian Noise BC}), uniform random noise (\textbf{Uniform Noise BC}), and estimate the noise covariance online (\textbf{DART} \cite{dart}). 
Results in Fig. \ref{fig:dubins_noise} show that alternative variants fail to improve the system safety, highlighting the importance of accounting for the safety criticality of a state during data augmentation. 
By using HJ reachability, SAFE-GIL is \textit{targetedly} guiding the agent towards safety-critical states, resulting in safer policies.

\noindent \textbf{\textit{Effect of disturbance bound.}} 
The guidance and learning of the safety-critical behavior in SAFE-GIL is dependent on the choice of the disturbance bound.
Thus, we perform an ablation study to understand the effect of the disturbance bound on the safety of the resultant policy (Table \ref{tab:dubins_d_bound}).
As the disturbance bound increases, SAFE-GIL is able to expose the demonstrator to higher potential learning errors (modeled as disturbance) and guide it towards more unsafe states.
Thus, a higher disturbance bound results in a more robust imitation policy.
However, increasing the disturbance bound beyond a particular point leads the system to states that are too risky, which the expert might fail to recover from. 
Moreover, as the training data shifts towards more unsafe states, the task performance of the learned policy decreases.
\vskip -0.1in
\begin{table}[h]
\resizebox{\columnwidth}{!}{%
\begin{tabular}{llll}
\hline
$\bar{d}_{max}$ & 0.2           & 0.4           & 0.6           \\ \hline
Collision Rate  & $0.19\pm0.02$ & $0.11\pm0.05$ & $0.07\pm0.01$ \\
Cost            & $2975\pm2$    & $3089\pm3$    & $3156\pm3$    \\ \hline
\end{tabular}%
}
\caption{Mean collision rates and cost of safe trajectories of agents trained with SAFE-GIL for different disturbance bounds.}
\label{tab:dubins_d_bound}
\end{table}
\vskip -0.1in

\begin{figure}[t]
\centering
\includegraphics[width=0.48\textwidth]{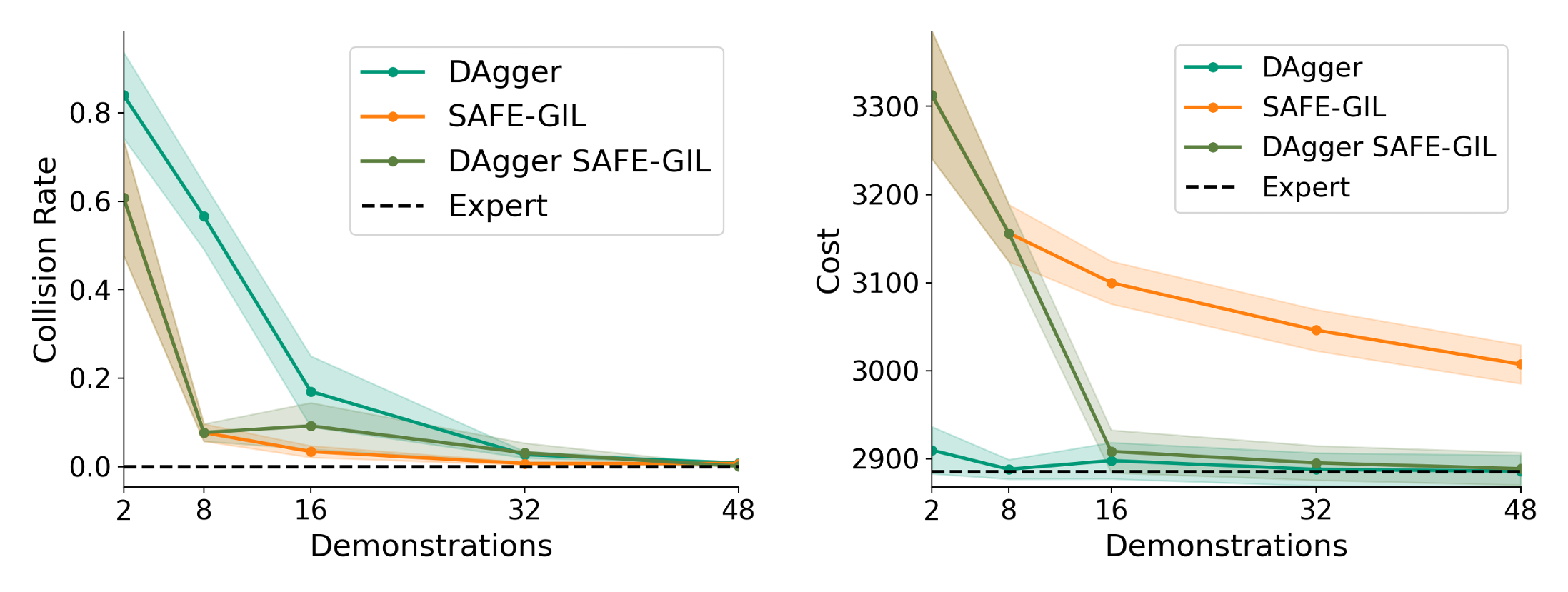}
\caption{SAFE-GIL can be combined with other imitation learning approaches to have complementary safety and performance advantages.}
\label{fig:dubins_dagger}
\vskip -0.2in
\end{figure}

\noindent \textbf{\textit{Mitigating Covariate Shift.}}
We demonstrate that SAFE-GIL can easily be integrated with existing approaches to reduce covariate shift and performance degradation, while mitigating safety-critical failures. As an example, we consider \textbf{DAgger} \cite{dagger} that mitigates covariate shift by iteratively aggregating expert actions on the states visited by the learned policy.
We combine DAgger with SAFE-GIL by injecting adversarial disturbance during expert rollouts. 
We see significant improvement in safety compared to DAgger alone and in performance compared to SAFE-GIL alone (Fig. \ref{fig:dubins_dagger}).

\subsection{Autonomous Aircraft Taxiing Using a Vision-Based Policy}
We now consider an aircraft taxiing task, used as a benchmark for robust perception and verification \cite{taxinet, taxinet2, taxinet3}. 
The agent is an aircraft simulated in an X-Plane flight simulator.
The system dynamics are modeled as: $\dot{p_x} = v \sin(\theta), \quad  \dot{p_y} = v \cos(\theta), \quad
\dot{\theta} = (v/h)\tan(\omega)$,
where $(p_x, p_y, \theta)$ denote the crosstrack error (CTE), the downtrack position (DTP), and the the heading error (HE) with respect to the runway centerline, respectively. 
The aircraft controls its steering $\omega$, bounded by $|\omega| \leq \bar{\omega}$.
In our example, we use $\bar{\omega} = 1 \, rad/s$, $v = 5 \, m/s$, and $h = 5 \, m$.
The task is to reach the end of the runway ($p_y=200$) without leaving the runway, which corresponds to $L = \{x: |{p}_x| \geq 10\}$. 
The expert is a PID controller designed to keep the aircraft on the runway and steer towards the centerline.
In this case, the expert has privileged access to the system state during the data collection procedure, which is \textit{not} available to the imitator during test time.
Instead, it needs to imitate the expert based on the RGB image observations obtained through a camera mounted on the plane's right wing.

\begin{figure}[h!]
\vspace{-1em}
\centering
\includegraphics[width=0.5\textwidth]{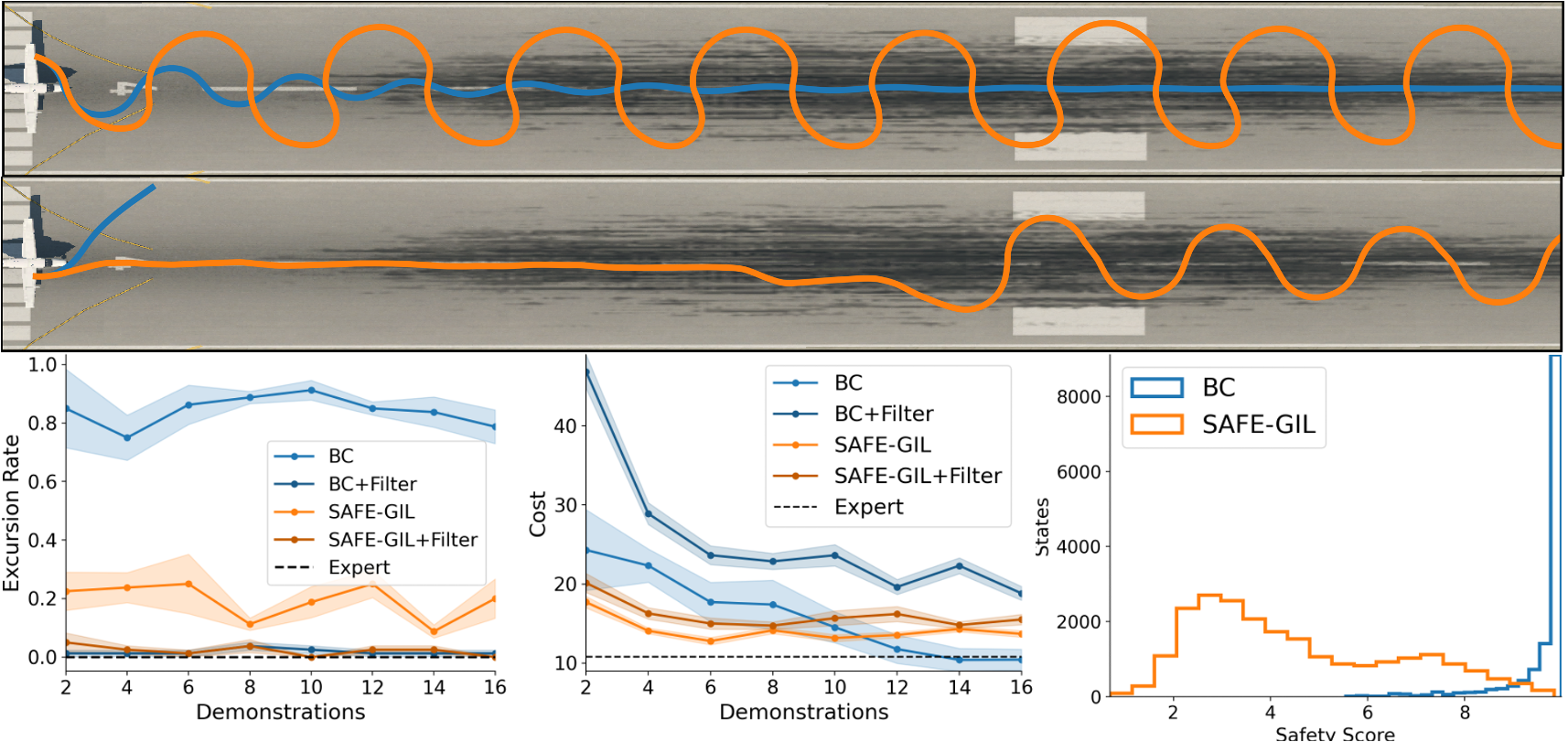}
\caption{Top: Expert demonstration with and without disturbance injection. Middle: BC and SAFE-GIL rollouts from the same initial state. BC fails to keep the aircraft on the runway. Bottom: Mean excursion rate (Left) and Mean squared distance from the centerline (Middle) vs number of demonstrations. Safety value distribution of the collected demonstrations (Right) is shifted towards lower values for SAFE-GIL.}
\label{fig:taxiing_main}
\vskip -0.1in
\end{figure}

We compute the value function and optimal disturbance using the Level Set Toolbox \cite{level_set_toolbox}.
We train each imitator for 5 different seeds to capture the performance variance and compare their \textit{excursion rates} (the percentage of trajectories that go out of the runway)  starting from 16 randomly sampled initial states.
We use mean squared distance from the centerline of the safe trajectories as the performance metric.
As evident from Fig. \ref{fig:taxiing_main}, SAFE-GIL achieves a substantially lower excursion rate compared to BC. 
This highlights the effectiveness of the safety guidance for learning robust imitation policies. 
During demonstrations, the disturbance guides the expert towards the runway boundaries. In turn, the agent is able to recover from policy errors that may steer the system towards the runway boundaries, whereas BC fails to do so.
This is also evident from the shift towards lower regimes in the safety value ($V^*(x)$) of the dataset collected with SAFE-GIL (Fig \ref{fig:taxiing_main}).
Overall, this case study demonstrates that even though SAFE-GIL relies on state-based guidance during data collection, it does not require any such privileged information during the test time and can readily be applied to observation-based policies.

\begin{figure}[h]
\vspace{-0.5em}
\centering
\includegraphics[width=0.48\textwidth]{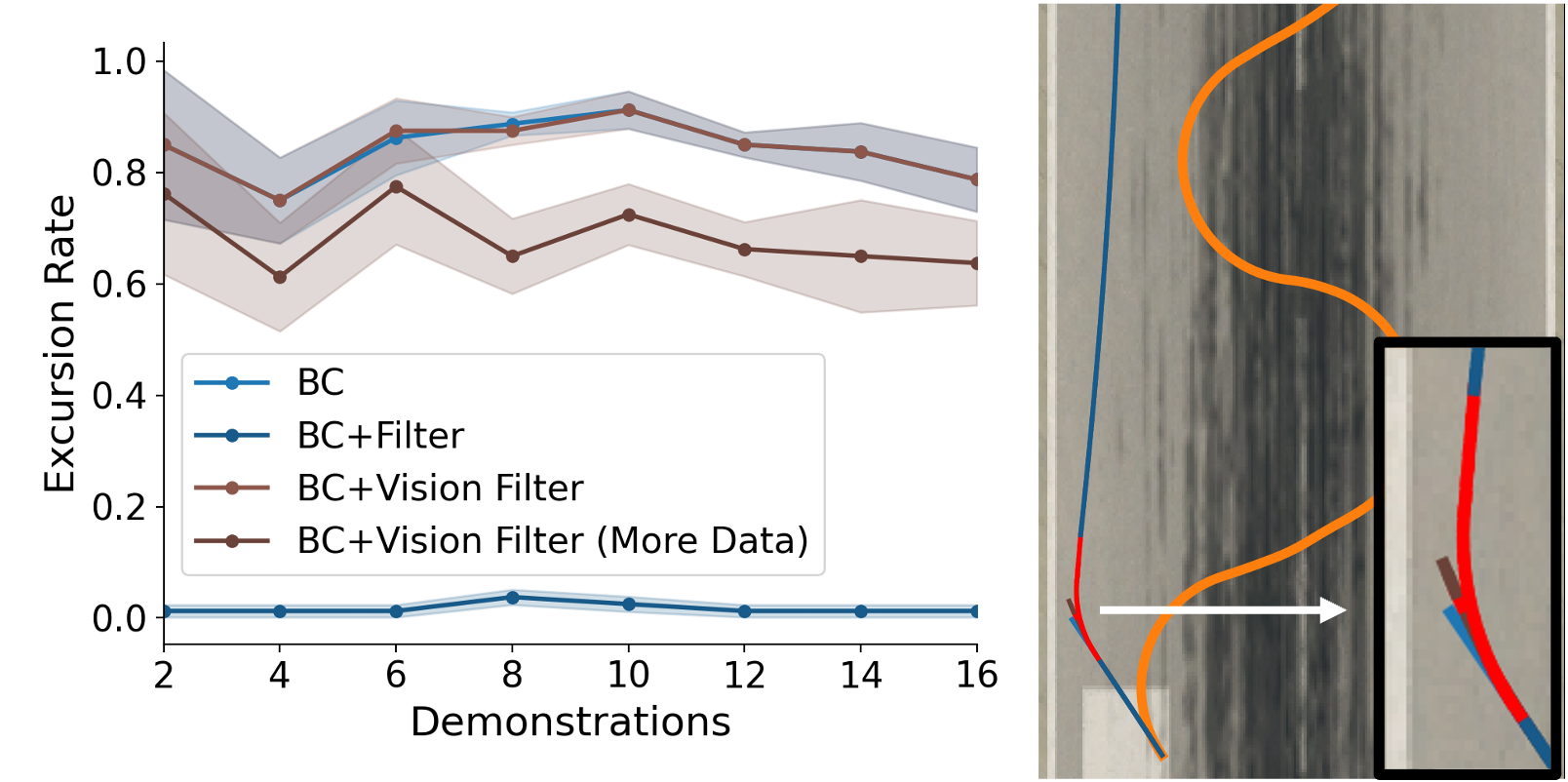}
\caption{Left: Mean excursion rates vs number of demonstrations. Right: SAFE-GIL, BC+Filter, BC+Vision Filter (More Data) trajectories from the same initial state. States where safety filter engages are denoted with red.}
\label{fig:taxiing_filtering}
\end{figure}

\noindent \textbf{\textit{Safety Filtering.}}
We now compare SAFE-GIL against a safety filter-based approach, a popular mechanism to ensure safety during test time \cite{wabersich2023data}.
We implement a least restrictive safety filter that overrides the imitation policy with a safe controller whenever the system approaches the BRT boundary \cite{borquez2023safety}. 
One of the key challenges in this example is the computation of the BRT for filtering since the state information is not available during the test time. Thus, we implement three different filters:
for \textbf{BC+Filter}, privileged state information is used during test time.
For \textbf{BC+Vision Filter} we train a state estimator from the images captured from states visited during expert demonstrations.
For \textbf{BC+Vision Filter (More Data)} we sampled 10000 (image, state) pairs uniformly over the state space to train the state estimator. During testing, a least restrictive filter is applied based on the estimated state of the agent.

The results are shown in Fig. \ref{fig:taxiing_filtering}.
While BC+Filter is able to maintain system safety, the overall performance of the system is still limited by the underlying policy. 
This can be seen in the trajectory shown in Fig. \ref{fig:taxiing_filtering}. 
Since the underlying policy doesn't have a good notion of recovery behavior, even though the filter keeps the agent on the runway, the agent isn't able to return to the centerline, whereas SAFE-GIL recovers much earlier and returns to the centerline. 
This highlights the need for \textit{design-time methods} that proactively reason about system safety during the training time itself.
Moreover, the performance of the safety filter degrades as we move away from the perfect state estimation assumption. 
BC+Vision Filter (More Data) with additional access to the environment data has only slightly lower excursion rates than BC, whereas the vision filter trained without any additional data performs similarly to BC. 
In Fig. \ref{fig:taxiing_filtering} (Right), we see that the vision filter with additional data engages safety control too little and too late due to errors in state estimation, highlighting the challenges associated with safety filtering 
when a system operates on raw sensory data.
Finally, we note that safety filtering, being a test-time method, is complementary to SAFE-GIL.
For instance, \textbf{SAFE-GIL+Filter} uses a safety filter on SAFE-GIL policies. Fig. \ref{fig:taxiing_main} shows that this further reduces the test-time failures without much degradation in task performance since the underlying policy is better able to reason about the safety of the system and recovery.

\noindent \textbf{\textit{Effect of adversarial disturbance}.}
Similar to the autonomous navigation case study in Section \ref{dubins_ex}, we compare SAFE-GIL against alternative noise injection methods ( Gaussian Noise BC, Uniform Noise BC, and DART).
We notice a similar pattern as before, highlighting the importance of the adversarial nature of injected disturbance for safety improvement. We omit those results due to space constraints.

\begin{table}[h!]
\resizebox{\columnwidth}{!}{%
\begin{tabular}{llll}
\hline
$\bar{d}_{max}$ & 0.1            & 0.3            & 0.5            \\ \hline
Excursion Rate  & $0.66\pm0.06$  & $0.63\pm0.08$  & $0.11\pm0.02$  \\
Cost            & $9.52 \pm 0.7$ & $12.6 \pm 0.6$ & $14.1 \pm 0.4$ \\ \hline
\end{tabular}%
}
\caption{Mean excursion rates and squared distance from the centerline for the agents trained with SAFE-GIL for different disturbance bounds.}
\vspace{-1em}
\label{tab:taxiing_d_bound}
\end{table}
\noindent \textbf{\textit{Effect of disturbance bound.}} 
As the disturbance bound increases, SAFE-GIL is able to expose the demonstrator to more unsafe states. As before, this results in better learning of the safety critical behavior at the expense of some task performance as seen in Table \ref{tab:taxiing_d_bound}.

\subsection{Quadrotor Navigation: Hardware experiment}
\label{quadrotor_hardware_example}
\vspace{-1em}
\begin{figure}[ht]
\centering
\includegraphics[width=0.5\textwidth]{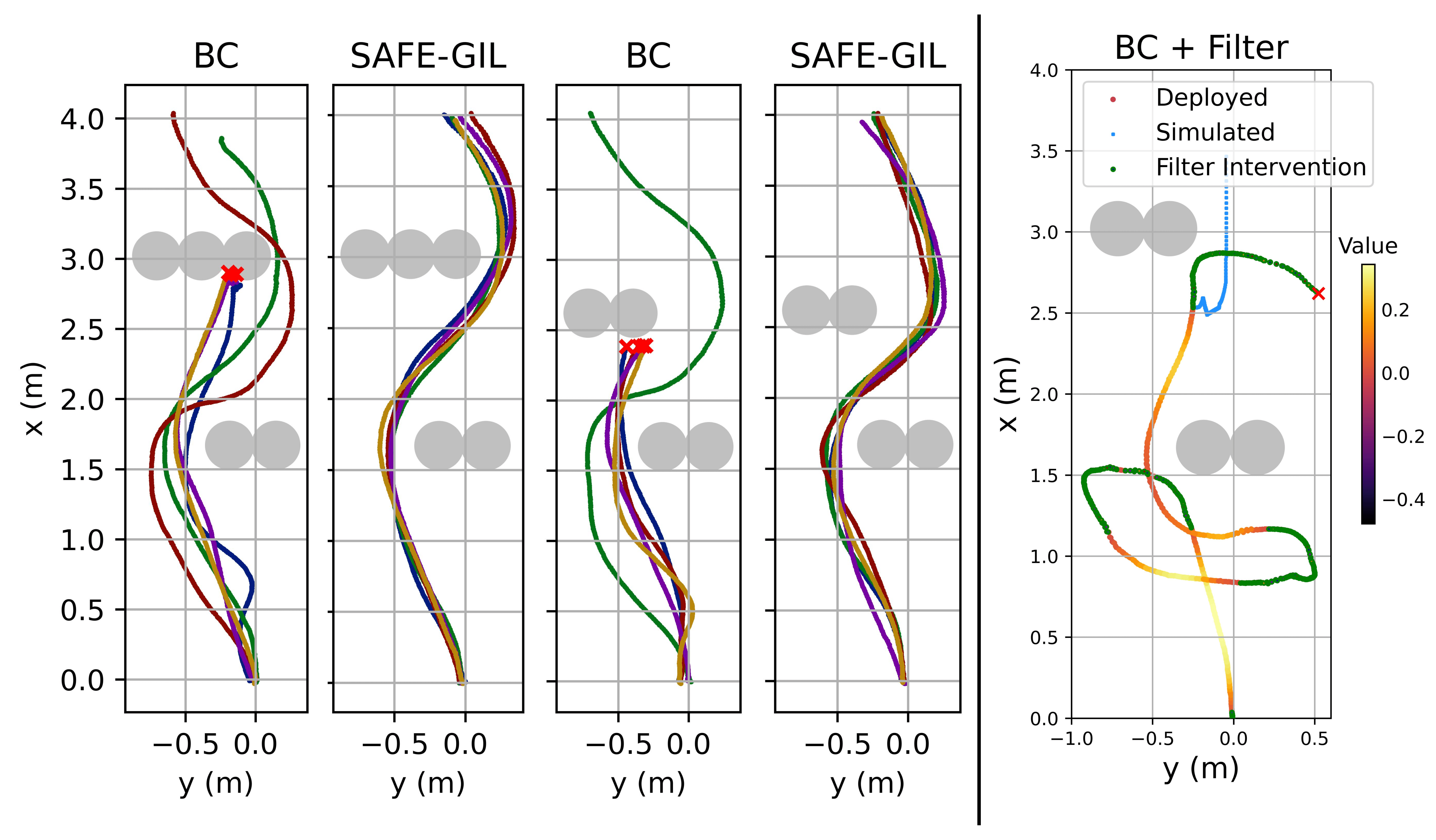}
\vspace{-2em}
\caption{Left: Trajectories of BC and SAFE-GIL in unseen settings, where X denotes a collision. SAFE-GIL leads to a significantly more robust performance compared to BC. Right: BC+Filtering trajectory.}
\vspace{-1em}
\label{fig:quadrotor_trajs}
\end{figure}

We now demonstrate our method on a Crazyflie 2.1 quadrotor testbed where the robot needs to reach a goal location without collisions.
In this case, we conduct experiments comparing BC, BC+Filter, and SAFE-GIL.
To collect the demonstrations, a \textit{human expert} uses a game controller to fly the Crazyflie and collect the applied roll and pitch commands (which are then tracked internally by a PID controller) along with estimated states ${p_x, p_y, p_z, v_x, v_y, v_z}$ ($xyz$ position and velocities, computed fully onboard using an optical flow camera), and an 8-pixel row of depth measurements for obstacles. 
A simplified quadrotor dynamics model is used for the disturbance computation in SAFE-GIL and safety filtering: $\dot{p}_x = v_x, \quad \dot{p}_y = v_y, \quad \dot{p}_z = v_z, \quad \dot{v}_x = a_g \tan u_\theta, \quad \dot{v}_y = -a_g \tan u_\phi, \quad \dot{v}_z = u_T - a_g$.
The control inputs are the net upwards thrust $u_T$ and desired roll $u_\phi$ and pitch $u_\theta$ angles.
The failure set includes all the cylinder obstacles (gray circles) and the x-y position outside the grid shown in Fig. \ref{fig:quadrotor_trajs}.
The optimal disturbance is computed by solving the HJI-VI through a DeepReach \cite{deepreach} neural network, which is then distilled for onboard computation.

\begin{table}[h]
\resizebox{\columnwidth}{!}{%
\begin{tabular}{cllll}
\hline
\multicolumn{1}{l}{Obstacle Configuration} & $(1,1)$ & $(2,2)$ & $(2,3)$ & $(2,2_c)$  \\ 
\hline
BC  & 20\% & 60\% & 60\%  & 80\% \\
\textbf{SAFE-GIL} & 0\% & 0\%& 0\%& 0\% \\         \hline     
\end{tabular}
}
\caption{Collision rates for different obstacle configurations. $(2,3)$ denotes two obstacles in the first row and three obstacles in the second row. Whereas, $2_c$ marks that the two obstacles rows are moved closer.}
\label{table:quad_collision_rate}
\vspace{-1em}
\end{table}
We collect 5 demonstrations on 2 different obstacle settings, first with singular obstacle pairs and another with dual obstacle pairs.  
The trained policies are tested five times each across four obstacle settings, two of which are not observed during the training time (Table \ref{table:quad_collision_rate}).
Trajectories for unseen obstacle settings are shown in Figure \ref{fig:quadrotor_trajs}, where one has an additional third obstacle, and the other has the second set of obstacles brought closer to the first to make the passage between them narrower.
SAFE-GIL outperforms BC across all configurations, achieving 0\% collision rate.
We find that despite using an approximate dynamics model, it is still useful in providing safety guidance through SAFE-GIL, highlighting that an exact dynamics model of the system may not be necessary for SAFE-GIL.

\noindent \textbf{\textit{Safety Filtering.}} 
We implement a least restrictive filter for BC by querying the value function (one for each obstacle setting) used during guidance. BC+Filter achieves a 0\% collision rate on obstacle setting $(1,1)$ and 40\% in $(2,2)$. 
Ideally the safety filter should keep the system safe at all times. 
However, the approximate dynamics model used for reachability computations assumes instantaneous direct control over the roll and pitch of the quadrotor, whereas delays between the desired and realized control during deployment result in failures.
To further test this, we propagate simulated approximate dynamics with no control delay in blue at the moment the safety filter intervenes (Fig. \ref{fig:quadrotor_trajs} (Right)). Indeed, the simulated filter maintains safety and the underlying policy is able to complete the task, highlighting the dependence of the safety filter on an accurate dynamics model.
Moreover, even when the safety filter keeps the agent safe, the performance of the filtered policy is still limited by the underlying BC policy. This is seen in Fig. \ref{fig:quadrotor_trajs} (Right). The filter protects the BC policy from a collision with the first set of obstacles early on. However, the BC policy isn't able to bring the agent back to the task when the safety filter disengages and the quadrotor swings from one side of the task space to the other until the safety filter engages again to keep the agent safe. 
\section{Discussion and Future Work} \label{sec:conclusion}
We propose SAFE-GIL, a new algorithm to learn safety-aware imitation policies. 
Our experiments show that SAFE-GIL results in significantly safer imitation policies, especially in low data regimes where the prediction error might be high.
However, the proposed framework has a few limitations:
 
\noindent \textit{Scalability of $d(x)$ computation}. 
Our approach requires a known dynamics model to solve for HJI-VI in order to compute $d(x)$.
Furthermore, HJI-VI is known to scale poorly for high-dimensional systems. To overcome these challenges, we will explore learning-based reachability methods \cite{deepreach, hsu2023isaacs}. These methods are shown to be scalable to complex high-dimensional systems, as well as to black-box dynamics models, and have already shown promise in our experiments.

\noindent \textit{Knowledge of the safety constraints}. The proposed approach relies on the known definitions of safety. An interesting future direction would be to learn the safety constraints from data, e.g., using inverse RL \cite{kim2024learning}.

\noindent \textit{Practicality of disturbance scale tuning}. The current approach is limited by an informed selection of disturbance bounds to maintain a reasonable safety-performance tradeoff. In our experiments, we have found 20-30\% disturbance error (compared to the control bounds) results in the best safety to performance ratio. Nevertheless, an interesting extension could be to estimate the disturbance bounds online based on the actual policy error.

\noindent \textit{Compensating for added disturbance during data collection.} The expert needs to compensate for the disturbance added during data collection, which can be challenging for high disturbance bounds.
One way to overcome this challenge could be to use a safety filter \textit{during training time} along with the proposed approach to compensate for any last-resort safety failures that the expert policy can't handle.




\section*{ACKNOWLEDGMENT}
This work is supported in part by the NSF CAREER Program under award 2240163 and the DARPA ANSR program.

\newpage
\bibliographystyle{IEEEtran}
\bibliography{Bib/reachability, Bib/safe_motion_planning, Bib/bansal_papers, Bib/cbf_and_clf, Bib/mpc_based_safety, Bib/sial_bib,  Bib/CDC2023, Bib/references}

\end{document}